\documentclass{opt2021} 

\title[Stochastic Learning Equation using Monotone Increasing Resolution of Quantization]{Stochastic Learning Equation using Monotone Increasing Resolution of Quantization}

\optauthor{%
\Name{Jinwuk Seok} \Email{jnwseok@etri.re.kr}\\
\Name{Jeong-Si Kim} \Email{sikim00@etri.re.kr}\\
\addr High Performance Embedded System SW Research Section, Future Computing Research Division, Artificial Intelligence Research Lab. Electronics and Telecommunications Research Institute, Rep. Korea}

\newtheorem{assumption}{Assumption}

\begin{document}

\maketitle

\begin{abstract}
In this paper, we propose a quantized learning equation with a monotone increasing resolution of quantization and stochastic analysis for the proposed algorithm. 
According to the white noise hypothesis for the quantization error with dense and uniform distribution, we can regard the quantization error as i.i.d.\ white noise. Based on this, we show that the learning equation with monotonically increasing quantization resolution converges weakly as the distribution viewpoint. 
The analysis of this paper shows that global optimization is possible for a domain that satisfies the Lipschitz condition instead of local convergence properties such as the Hessian constraint of the objective function. 
\end{abstract}

\section{Introduction}
A lot of researchers have regarded quantization as an efficient methodology enabling signal processing by reducing the amount of computation in small-scale hardware in the field of engineering(\citet{Boutalis_1989, Weiss_1979, Ljung_1985}, including machine learning (\citet{NIPS2016_6504, NIPS2015_5784, DBLP:conf/iclr/HanPNMGTEVPTCD17}).
However, due to error generation and propagation by quantization, designing quantization in the signal processing has been updating the least significant bits corresponding directional derivative (\citet{seide2014-bit, NIPS2017_6768, journals/corr/HubaraCSEB16}).  
This type of quantization is vulnerable in local minima, occurring degradation of performance in signal processing, and the same problem occurs in the learning equation to machine learning.
In this paper, we provide a novel quantized learning equation that can find an optimal point in a domain satisfying Lipschitz continuous to be robust to local minima.  
Increasing the resolution of quantization with respect to time, we note that the learning equation can find the global optimal point within such a domain by stochastic analysis.  
Numerical experiments show that the proposed methodology overcomes the weak point in conventional quantization, such as degradation of performance caused by quantization.  


\section{Fundamental Definition and Formulation for the Proposed Algorithm}
We set the following definitions and assumptions, before beginning of our discussion.
\begin{definition}
For $x \in \mathbf{R}$, we define the quantization as follows:
\begin{equation}
    x^Q \triangleq \frac{1}{Q_p} \lfloor Q_p \cdot  (x + 0.5 \cdot Q_p^{-1}) \rfloor
= \frac{1}{Q_p} \left( Q_p \cdot x + \varepsilon \right) = x + \varepsilon Q_p^{-1}, \quad x^Q \in \mathbf{Q}.
\label{define_eq01}
\end{equation}
, where $x^Q \in \mathbf{Z}$ is an integral part of $x \in \mathbf{R}$, $Q_p \in \mathbf{Z}^{+}$ is a quantization parameter, and $\varepsilon$ is a quantization error such as $\varepsilon \in \mathbf{R}[-0.5, 0.5]$.  
\label{def_01}
\end{definition}

\begin{assumption}
\label{assum01}
For $x_t \in B^o (x^*, \rho)$ ,  there exist a positive value $L$ w.r.t. a scalar field $f(x) : \mathbf{R}^n \rightarrow \mathbf{R}$ such that
\begin{equation}
\| f(x_t) - f(x^*) \| \leq L \| x_t - x^* \|, \quad \forall t > t_0
\label{Ass01_eq01}    
\end{equation}
where $B^o (x^* , \rho)$ is an open ball such that $B^o (x^* , \rho) = \{ x | \| x - x^* \| < \rho \}$, and $f(\cdot)$ is an objective function. 
\end{assumption}

In $\eqref{define_eq01}$, we replace the constant quantization parameter $Q_p$ with a monotone increasing quantization parameter concerning time $t$ such as $Q_p(t)$. Thereby, we obtain the quantization error term as a monotone decreasing function for time $t$. 
If a quantization error exists densely distributed and follows a uniform distribution,  we can regard the quantization error to be a white noise according to \citet{Gray:2006, DBLP:journals/tsp/BarnesTL85, Sripad:1977, Claasen:1981}. Additionally, in case of which a quantization error is a vector such as  $\vec{\varepsilon} \in \mathbf{R}^n$, if it is pairwise independent and follows a uniform distribution asymptotically, \citet{Jimenez_2007} proves that the vector valued quantization error is a white noise as well.   
Therefore, we regard the quantization error vector as a white noise, without proof. 

When the weight vector $w_{t} \in \mathbf{R}^n, \; w_{t} = \{w_t^1, w_t^2, \cdots w_t^n\}$ and a learning rate $\lambda_t = \alpha \in \mathbf{R}(0, 1), \; \forall t \in \mathbf{Z}^+$ are given, , we can obtain a canonical formulation for quantized learning equation as follows:
\begin{equation}
w_{t+1} = w_t - \lambda_t \cdot h(w_t) 
\label{stochatic_eq01}    
\end{equation}
, where $h(w_t)$ is a directional derivative corresponding to the objective function $f(w_t, x_t)$ for machine learning such that $h(w_s) \triangleq (J \circ \nabla f)(w_s)$ for some function $J$ .  For example,  if $J(w_s) $ is an identity function such that $J \circ f (x) = f(x)$, $h(w_s) = \nabla f(w_s)$ , where $f(w_s)$ is an objective function for a machine learning algorithm.  

Suppose that the parameter vector of the current step $w_t$ and the next step $w_{t+1}$ are quantized, we have 
\begin{equation}
w_{t+1}^Q = \left(w_t^Q - \lambda_t \cdot h(w_t) \right)^Q 
= w_t^Q - \left(\lambda_t \cdot h(w_t) \right)^Q
\label{stochatic_eq02}    
\end{equation}
, where the $\vec{\varepsilon}$ is the vector valued quantization error so that the distribution of components are independent distribution defined $\vec{\varepsilon} \in \mathbf{R}^n$. 
If there exist a rational number $\alpha_t \in \mathbf{Q}(0, Q_p)$ instead of $\lambda_t$ , we have the following quantized learning equation by simple calculation.
\begin{equation}
w_{t+1}^Q 
= w_t^Q - \frac{\alpha_t}{Q_p} \cdot Q_p h(w_t) + {\varepsilon_t} Q_p^{-1} 
= w_t^Q - \frac{\alpha_t}{Q_p} (Q_p h(w_t))^Q \;\;\; \because \alpha_t \in \mathbf{Q}.
\label{stochatic_eq04}
\end{equation}
Hereby, we can obtain the search equation providing the quantized parameter vector for all steps $t \in \mathbf{N}$ by the mathematical induction.

\section{Stochastic Analysis}

\subsection{Analysis of the Proposed Quantization}

If the quantization error vector ${\vec{\varepsilon}}_t \in \mathbf{R}^n$ satisfying the WNH, \citet{klebaner, Gray:2006} shows that the deviation can be calculated as follows:
\begin{equation}
\forall \vec{\varepsilon}_t \in \mathbf{R}^n,
\mathbb{E} Q_p^{-2} \vec{\varepsilon}^2_t = \mathbb{E}  Q_p^{-2} \cdot tr(\vec{\varepsilon}_t \vec{\varepsilon}^T_t) = \frac{1}{12 \cdot Q_p^2} \cdot n. 
\label{stochatic_eq06}    
\end{equation}
Since the WNH establishes that the quantization error is a i.i.d.\ white noise, we can regard that weight vector $w_t^Q \in \mathbf{R}^n$  as a stochastic process $\{ W_t \}_{t=0}^{\infty}$.
Suppose that $N$ is the number of needed data while the weight vector is updated. In other words, if $t$ is an index of epoch, then $N$ is total number of  data, and if $t$ is an index of mini-batch, $N$ is the number of data in a unit mini-batch. 
Let a granular time index  $s:\mathbf{Z}[0, N) \rightarrow \mathbf{R}[0, 1), \;s \in \mathbf{R}[t, t+1)$ such that $s (\tau) = \frac{1}{N} \tau$ , where $\tau \in \mathbf{R}[0, N)$ so to set the time index between $t$ and $t+1$.   Additionally,  we replace time index such that  $t=t_1$ and $t+1 = t_N$, for convenience.
Let $Z(s(\tau)) = w_{t_1} + s(\tau)(w_{t_N} - w_{t_1})$.  By chain rule, we can obtain
\begin{equation}
\int_0^1 dZ(s) 
= \int_0^1 \frac{\partial Z(s)}{\partial s} ds 
= \int_{t}^{t+1} dw_{s} 
= \int_0^1 \left(-\alpha_t \nabla f(w_{t_1}) + \vec{\varepsilon_s} Q_p^{-1}(t_1) \right) ds. 
\label{th02-pf01}    
\end{equation}
Differentiate both sides in $\eqref{th02-pf01}$ to $s$.  Additionally, letting $\vec{\varepsilon_s} ds = \sqrt{\frac{n}{12}} dB_s $ from $\eqref{stochatic_eq06}$ and WHN,  we get the following stochastic differential equation to the proposed learning equation:
\begin{equation}
dW_s = -\alpha_{t} h(W_{t}) ds + \vec{\varepsilon_s} Q_p^{-1}(t) ds =  -\alpha_{t} h(W_{t}) ds + \sqrt{\frac{n}{12}} Q_p^{-1}(s) d\vec{B_s}
\label{th02-pf05}    
\end{equation}
where $dB_s$ is the differential of a vector valued standard Wiener process with mean zero and variance one.   

\begin{theorem}
If the stochastic differential equation induced by the learning equation $\eqref{stochatic_eq04}$ satisfies $\eqref{th02-pf05}$, the stochastic process $\{ W_t \}_{t=0}^{\infty}$ generated by the learning equation weakly converges to the global minimum on the domain defined in Assumption $\ref{assum01}$ when the deviation of the quantization error is given as follows:
\begin{equation}
\inf_{t \geq 0} \sigma(t) = \frac{C}{\log (t + 2)}, \quad C \in \mathbf{R},\; C \gg 0
\label{stochatic_eq14}    
\end{equation}
where, $\sigma(t) = \sqrt{\frac{n}{24}}Q_p(s)^{-1}$. 
\label{theorem-1}
\end{theorem}
Theorem $\ref{theorem-1}$ means that if we properly increase the quantization resolution $Q(s)$, the proposed quantization learning equation can find the global minima of an objective function within the domain satisfying Lipschitz continuous is satisfied. 
We provide a detailed proof of Theorem in Appendix.

\subsection{Scheduler function}
\begin{figure*}[!t]
\centering
\subfigure[]{
\includegraphics[width=.45\columnwidth]{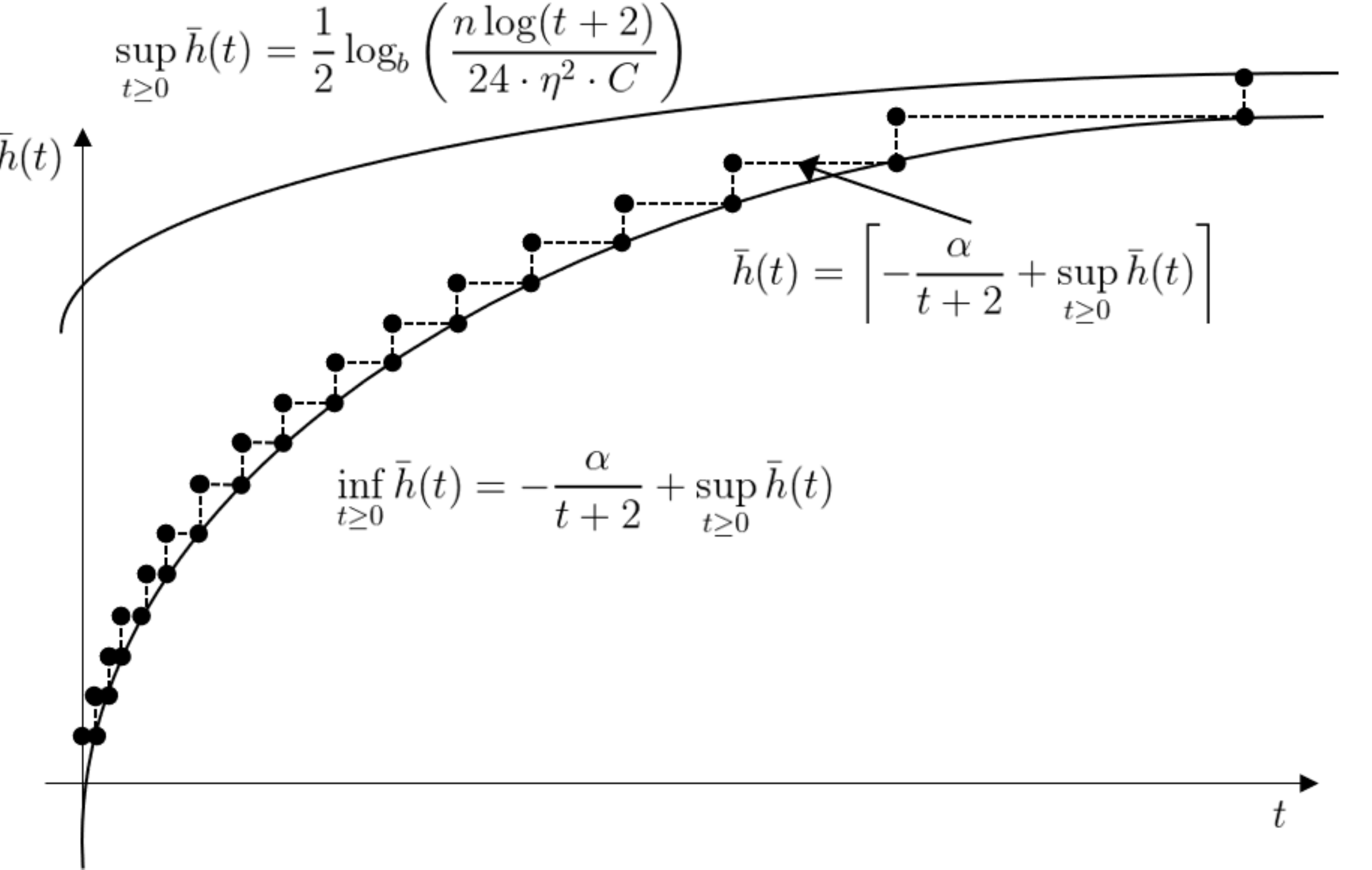}
\label{power-case}
}
\centering
\subfigure[]{
\includegraphics[width=.45\columnwidth]{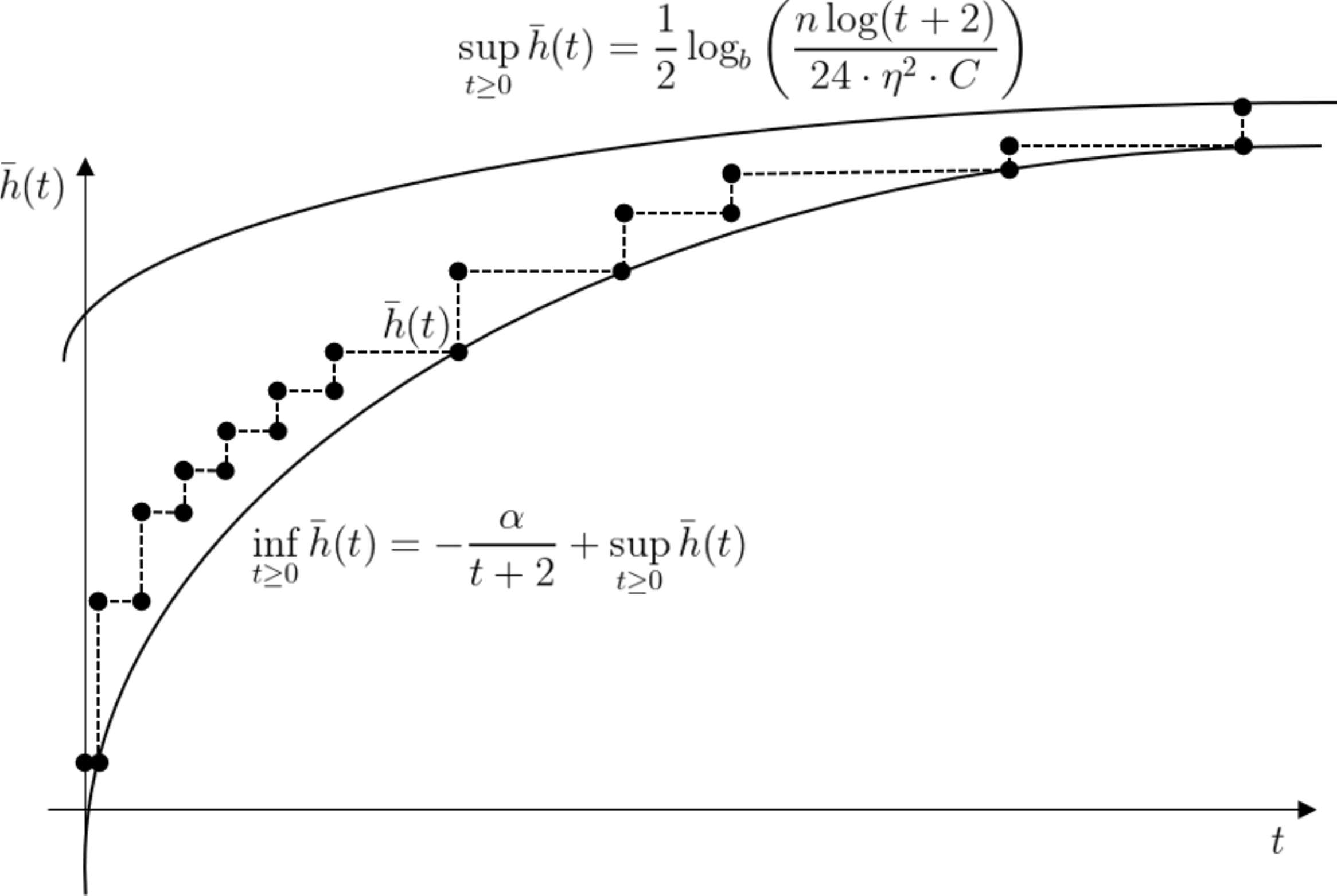}
\label{practical-case}
}
\caption{
(a) Theoretical trend of $\sigma(t)$ based on the infimum
(b) Practical trend of $\sigma(t)$ to avoid the vanishing gradient by the significant figure owing to quantization
}
\label{Fig-01}
\end{figure*}

Since $\sigma(t) \in \mathbf{R}$ is a proportional value to $Q_p(t) \in \mathbf{Z}$, we can't apply the result of Theorem $\ref{theorem-1}$ to the proposed quantization.  
However, if there exists a feasible $\sigma(t) \in \mathbf{Z}$ such that $\sigma(t) \geq \inf \sigma(t) \triangleq c/\log (2+t)$, it satisfies the Theorem $\ref{theorem-1}$.
Furthermore, if there exists the supremum of $\sigma(t)$ such that  $\inf \sigma(t) \leq \sigma(t) \leq T(t)$, the proposed quantization that satisfies the condition for global optimization is possible avoiding a extreme 1-bit quantization at early stage. 
For this, we define the quantization parameter $Q_p(t)$ depending on a monotone increasing function $\bar{h}(t) \in \mathbf{Z}^+$ with respect to time t. 
\begin{equation}
Q_p (t) = \eta \cdot b^{\bar{h}(t)}, \quad \text{such that } \;\; \bar{h}(t) \uparrow \infty \; \text{ as } \; t \rightarrow \infty.
\label{stochatic_eq17}    
\end{equation}
By a simple calculation using the results in the previous section, $\bar{h}(t)$ satisfying $\eqref{stochatic_eq17}$ has the following supremum and the infimum.
\begin{equation}
\frac{1}{2} \log_b \left( \frac{n}{24 \cdot \eta^2} \cdot T(t)^{-1} \right)
\leq \bar{h}(t) \leq 
\frac{1}{2} \log_b \left( \frac{n \log (t+2)}{24 \cdot \eta^2 \cdot C} \right)
\label{stochatic_eq18}    
\end{equation}
To specify the $\bar{h}(t)$, we let $T(t)$ for $b > 1$ such that $T(t) = b^{ \left( \frac{2\beta}{t+2} \right)} \cdot \inf_{t \geq 0} \sigma(t)$. Based on such $T(t)$, evaluating the supremum and the infimum of $\bar{h}(t)$, we can get the $\bar{h}(t)$ as follows:  
\begin{equation}
\bar{h}(t) 
\geq \frac{1}{2} \log_b \left( \frac{n}{24 \cdot \eta^2} \cdot T(t)^{-1} \right)
= -\frac{\beta}{t+2} + \sup_{t \geq 0} \bar{h}(t), \quad \because \sup_{t \geq 0} \bar{h}(t) = \frac{1}{2} \log_b \left( \frac{n \log (t+2)}{24 \cdot \eta^2 \cdot C} \right)
\label{stochatic_eq21}    
\end{equation}
Using $\eqref{stochatic_eq21}$, calculating the quantization parameter, we can obtain the $Q_p(t) = \eta \cdot b^{\bar{h}(t)}$ satisfying the result of Theorem $\ref{theorem-1}$. 
Figure $\ref{Fig-01}$ presents the conceptual diagram of the proposed schedule function.

\section{Numerical Experiments}
We conduct numerical experiments on the image classification problem to verify the empirical validation of the proposed algorithm and the analysis. 
The data set we employ in the experiment is the well-known CIFAR-10 data. 
The test network for the experiment is a ResNet with 32 Layers.  
The number of total samples to training is 50000, the testing data is 10000. 
We use the cross-entropy loss as the objective function provided by the PyTorch that is an A. I. framework based on python. 
We perform the ten training times with every 100 epochs, and we yield the classification accuracy from the average of Top-1 accuracy to the training and testing data set.
The algorithms used in the experiments are the general stochastic gradient descent(SGD) algorithm and the ADAM (ADaptive Moment Estimation) algorithm widely used in machine learning. 
We compared the data classification accuracy by combining the proposed quantization algorithm with each conventional algorithm.

Furthermore, in the proposed quantization, we set the quantization parameter to be $Q_p(0) = 2^{2}$ at an initial stage, and the scheduler function calculates the quantization parameter every epoch. 
In the result of the numerical experiments, the proposed algorithm shows better classification performance than both ADAM and SGD's, as represented in Table $\ref{table-01}$.
We can regard the result of the conventional quantization learning equation using only the lower 1 bit as the performance of the existing algorithm when it has the lowest learning rate in Table $\ref{table-01}$. 
In particular, when applied to ADAM, the performance improvement is higher than that of SGD.
We consider that such a result is based on ADAM's feasible search domain is wider than SGD's.

\begin{table}[tb]
\begin{tabular}{lllllllll}
\hline
              & \multicolumn{2}{l}{ADAM} & \multicolumn{2}{l}{QtADAM} & \multicolumn{2}{l}{SGD} & \multicolumn{2}{l}{QSGD} \\ \hline
Learning rate & Training    & Test      & Training  & Test      & Training  & Test      & Training& Test    \\ \hline
0.25          & 78.73	    & 69.26     & 78.16     & 71.28     & 95.78     & 77.27     & 95.95   & 77.27     \\
0.125         & 79.47	    & 69.00	    & 83.86     & 72.75     & 93.86     & 74.42     & 94.89   & 75.69     \\
0.0625        & 88.23	    & 74.46	    & 89.12     & 76.63     & 92.34     & 73.42     & 91.63   & 73.21     \\
0.03125       & 93.91	    & 78.91	    & 94.62     & 79.13     & 86.45     & 68.65     & 86.06   & 67.36     \\
0.015625      & 95.62	    & 80.24	    & 95.57     & 80.54     & 77.84     & 65.73     & 78.00   & 66.41     \\
0.0078125     & 95.95	    & 80.77	    & 96.78     & 81.39     & 71.24     & 65.74     & 70.10   & 65.01     \\
0.00390625    & 96.23	    & 81.20	    & 96.43     & 81.74     & 60.47     & 57.56     & 61.17   & 59.07     \\
0.001953125   & 96.13	    & 80.99	    & 95.77     & 79.59     & 50.59     & 49.73     & 50.17   & 48.83     \\
0.0009765625  & 95.71	    & 78.28	    & 94.84     & 77.12     & 43.49     & 42.66     & 44.18   & 43.19     \\ \hline
Average       & 91.11       & 77.01     & 91.68     & 77.79     & 74.67     & 63.91     & 74.68   & 64.00     \\ \hline  
\end{tabular}
\caption{
    The results of the numerical experiments about the classification of CIFAR-10 data set with various learning rates
}
\label{table-01}
\end{table}

\begin{algorithm2e}[tb]
    \caption{Learning equation with the proposed quantization scheme}
    \label{algorithm-1}
    \DontPrintSemicolon
    \SetAlgoLined
    \KwData{Data-set needed classification such as the CIFAR-10}
    \KwResult{Learned Network for Input Data}
    initialization\; 
    Set Parameters as $n \in \mathbf{Z}, \eta=1, b=2, \alpha \in \mathbf{Q}(0, 1]$ \;
    $t \leftarrow 0$ and Compute $\bar{h}(0)$ and $Q_p(0)$\; 
    \While{meet stopping criterion}{
        $h_t \leftarrow (J \circ \nabla f)(w_t)$ \tcp*{Common weight update}
        $h^Q \leftarrow \frac{1}{Q_p}(Q_p \cdot h_t)^Q$ \tcp*{Quantization}
        Avoid Gradient Vanishing and Check the Limit of $\bar{h}_t$ \tcp*{Quantization}
        Check the bound of $\bar{h}(t)$ \tcp*{Quantization}
        $w_t \leftarrow w_t -\alpha \cdot h_t^Q$         \tcp*{Common weight update}
        $t \leftarrow t+1$  \tcp*{to unit mini-batch or an epoch}
    }
\end{algorithm2e}

\section{Conclusion}
We present a quantization learning algorithm that monotonically increases the quantization resolution with respect to time and stochastic analysis of the proposed algorithm.  
The stochastic analysis of the proposed algorithm shows that the proposed quantization methodology can find the global optimum under the input domain satisfying Lipschitz continuous without any convex condition such as the limitation of Hessian. 
Therefore, we expect that the better the search capability of the conventional learning equation, the proposed algorithm shows better performance.
We verify that the proposed algorithm shows superior classification performance without any degradation by quantization from the result of the numerical experiments.

\newpage
\clearpage
\appendix
\section{Acknowledgement}
This work was supported by Institute for Information \& communications Technology Planning \& Evaluation(IITP) grant funded by the Korea government(MSIT) (No.2017-0-00142, Development of Acceleration SW Platform Technology for On-device Intelligent Information Processing in Smart Devices)

\section{Supplementary Information}
We provide the proof of the Theorem $\ref{theorem-1}$, the detailed procedure of the sub-functions for the proposed algorithm, and the information of hyper-parameters to each algorithm employed in numerical experiments.


\subsection{Proof of Theorem $\ref{theorem-1}$}

\begin{proof}
For the proof of the theorem, we depend on the lemmas in works of \citet{pub.1062843904}.
The aim of this section is to prove the following convergence of the transition probability:
\begin{equation}
\lim_{\tau \rightarrow \infty} \sup_{w_t, w_{t+\tau} \in \mathbf{R}^n} \| p(t, \bar{w}_t, t + \tau,  w^*) - p(t, w_t, t + \tau,  w^*) \| = 0
\label{lmpf2.7_eq01}    
\end{equation}
, where $t$ and $\tau$ is the epoch index and the iteration to a single data index, respectively. $w^*$ represents an optimal weight vector.

Let the infimum of the transition probability from $t$ to $t+1$ such that
\begin{equation}
\delta_t = \inf_{x, y \in \mathbf{R}^n} p(t, x, t+1, y) 
\label{lmpf2.7_eq02}    
\end{equation}

By the lemma in \citet{pub.1062843904}, the upper bound of $\eqref{lmpf2.7_eq01}$ is 
\begin{equation}
\overline{\lim_{\tau \rightarrow \infty}} \sup_{w_t, w_{t+\tau} \in \mathbf{R}^n} \| p(t, \bar{w}_t, t + \tau,  w^*) - p(t, w_t, t + \tau,  w^*) \| \leq 2 \| w^* \|_{\infty} \prod_{k=0}^\infty (1 - \delta_{t+k}).
\label{lmpf2.7_eq03}    
\end{equation}

Since $\prod_{k=0}^{\infty} (1 - \delta_{t+k}) \leq \exp(-\sum_{k=0}^{\infty} \delta_{t+k})$ ,  we rewrite $\eqref{lmpf2.7_eq03}$ as follows:
\begin{equation}
\overline{\lim_{\tau \rightarrow \infty}} \sup_{w_t, w_{t+\tau} \in \mathbf{R}^n} \| p(t, \bar{w}_t, t + \tau,  w^*) - p(t, w_t, t + \tau,  w^*) \| \leq 2 \| w^* \|_{\infty} \exp(-\sum_{k=0}^{\infty} \delta_{t+k}) ).
\label{lmpf2.7_eq04}    
\end{equation}

Herein, to obtain the bound of $\delta_{t+k}$, we rewrite the stochastic differential form derived from Theorem 2 as follows:
\begin{equation}
dW_s = - \nabla H(W_{s}) ds + \sigma(s) \sqrt{G} dB_s, \quad s \in \mathbf{R}(t, t+1).
\label{lmpf2.7_eq05}    
\end{equation}
, where $\sigma(s) \triangleq Q_p^{-1}(s)$ , $G = \frac{n}{12}$, and $\nabla H(W_{s}) = \lambda_s h(W_s) \triangleq \lambda_s \cdot (J \circ \nabla f)(W_s)$ for a function $J$ such that $\nabla H(W^*) = \lambda_{*} h(W^*) = (J \circ \nabla f)(W^*) = 0$ as represented in $\eqref{stochatic_eq01}$.  

Define a domain $\mathcal{F} \{ f: [t, t+1] \rightarrow \mathbf{R}^n, f \text{ continuous } \}$, 
Let $P_x$ be the probability measures on $\mathcal{F}$ induced by $\eqref {lmpf2.7_eq05}$ and $Q_x$ derived by the following equation:
\begin{equation}
d\bar{W}_{\tau} = \sigma(\tau) \sqrt{G} dB_{\tau}, \quad \tau \in \mathbf{R}(t, t+1). 
\label{lmpf2.7_eq06}    
\end{equation}

By the Girsanov theorem (introduced in \cite{klebaner, oksendal}), we obtain
\begin{equation}
\frac{dP_{w}}{dQ_{w}}
= \exp \left\{\int_t^{t+1} \frac{G^{-1}}{\sigma^2 (\tau)} \langle - \nabla H(W_{\tau}), d\bar{W}_{\tau} \rangle -\frac{1}{2} \int_t^{t+1} \frac{G^{-1}}{\sigma^2 (\tau)} \|  \nabla H(W_{\tau}) \|^2 d\tau \right\}.
\label{lmpf2.7_eq07}    
\end{equation}

To compute the upper bound of $\eqref{lmpf2.7_eq07}$, we will check the upper bound of $\| \nabla H \|$.  
Whereas, since $\|G\|$ is not depending on time index $s$,  we regard it as a constant value for all $s$. 
By definition, since the objective function is continuous, the gradient of $H(w_s)$ fulfills the Lipschitz continuous condition $\eqref{Ass01_eq01}$ too. 

Thereby,  for $w_t \in B^o (w^*, \rho)$ , there exist a positive value $L'$  such that 
\begin{equation}
\| \nabla f(w_{\tau}) - \nabla f(w^*) \| \leq L' \| w_{\tau} - w^* \|, \quad \forall \tau > 0.
\label{lmpf2.7_eq09}    
\end{equation}

Successively, by the definition of $\nabla H(W^*)$ being equal to zero,  the Lipschitz condition forms simply as follows : 
\begin{equation}
\| \nabla H(W_t) \| \leq L' \lambda_t \rho = C_0.
\label{lmpf2.7_eq10}    
\end{equation}

Consequently, for all $s \in \mathbf{R}[t, t+1)$, we compute the upper bound of the first term in exponential function as follows:
\begin{equation}
\begin{aligned}
&\left\| \int_t^{t+1} \frac{G^{-1}}{\sigma^2 (s) } \langle \nabla H(W_{s}), d \bar{W}_{s} \rangle \right\| 
\leq  \int_t^{t+1} \left\| \frac{G^{-1}}{\sigma^2 (s)} \langle \nabla H(W_{s}), d \bar{W}_{s} \rangle \right\| \\
&\leq  \int_t^{t+1} \frac{\left\| G^{-1}\right\|}{\sigma^2 (s)} \left\| \nabla H(W_{s}) \right\| \sigma (s) \sqrt{\|G\|} d B_{s}    
\leq  \frac{\sqrt{\left\| G^{-1}\right\|}}{\sigma (s)} \sup \left\| \nabla H(W_{s}) \right\|  \int_{t}^{t+1}  dB_s \\
&\leq  \frac{\sqrt{\left\| G^{-1}\right\|}}{\sigma (s)} C_0  \| B_t - \frac{1}{2} \|  
\leq \frac{1}{\sigma (s)} C_0 \sqrt{\left\| G^{-1}\right\|} (\rho + \frac{1}{2}). 
\end{aligned}
\label{lmpf2.7_eq14}    
\end{equation}

It implies that 
\begin{equation}
\left\| \int_t^{t+1} \frac{G^{-1}}{\sigma (s) } \langle - \nabla H(W_{\tau}, X_{\tau}), d \bar{W}_{\tau} \rangle \right\| \leq \frac{C_1}{\sigma (s)}
\label{lmpf2.7_eq15}    
\end{equation}
, where $C_1$ is positive value such that $C_1 > C_0 \sqrt{\left\| G^{-1}\right\|} (\rho + \frac{1}{2}) $.

In addition, the upper bound of the second term is  
\begin{equation}
\begin{aligned}
&\frac{1}{2} \left\| \int_t^{t+1} \frac{G^{-1}}{\sigma^2 (s) } \|  \nabla H(W_{s}) \|^2 d\tau \right\|  
\leq \frac{1}{2} \int_t^{t+1} \frac{\| G^{-1} \| }{\sigma^2 (s)} \|  \nabla H(W_{s}) \|^2 d\tau \\
&\leq \frac{1}{2} \frac{\| G^{-1} \| }{\sigma^2 (s)} \sup \|  \nabla H(W_{s}) \|^2 \int_t^{t+1} d\tau 
\leq \frac{1}{2 \sigma^2 (s)} \| G^{-1} \| \cdot C_0^2 
\leq \frac{C_2}{2 \sigma^2 (s)},   \quad \because C_2 > \| G^{-1} \| \cdot C_0^2. 
\end{aligned}
\label{lmpf2.7_eq16}
\end{equation}

By assumption, since $\sigma(s)$ is monotone decreasing function, the supremum of  $\sigma(s)$ is $\sigma(0) $ for all $s \in \mathbf{R}[0, \infty)$, i.e. $\sup_{s \in \mathbf{R}[0, \infty]}  \sigma(s) = s(0)$.
With the supremum of each term in $\eqref{lmpf2.7_eq07}$, we can obtain the lower bound of the Radon-Nykodym derivative $\eqref{lmpf2.7_eq07}$  such that
\begin{equation}
\frac{dP_{w}}{dQ_{w}} \geq \exp \left( - \frac{1}{\sigma (s)}\left( C_1 + \frac{C_2}{2 \sigma (s)} \right)\right) \geq \exp \left(- \frac{C_3}{\sigma (s)}  \right), \quad \because C_3 > 2 \sigma(0) C_2 + C_1.
\label{lmpf2.7_eq17}    
\end{equation}

Consequently, for any $\varepsilon > 0$  and $w_t, \; w^* \in \mathbf{R}^n$,  the infimum of $P_w (|W_{t+1} - w^*| < \varepsilon) $ is 
\begin{equation}
P_w (|W_{t+1} - w^*| < \varepsilon) 
\geq \exp\left(- \frac{C_3}{\sigma(s)} \right) Q_w (|W_{t+1} - w^*| < \varepsilon).
\label{lmpf2.7_eq18}    
\end{equation}

Since $Q_w$ is a normal distribution based on $\eqref{lmpf2.7_eq06}$,  we have
\begin{equation}
\begin{aligned}
&P_w (|W_{t+1} - w^*| < \varepsilon) 
\geq \exp\left(- \frac{C_3}{\sigma (s)} \right) \int_{\| x - w^* \| < \varepsilon} \frac{1}{\sigma(s) \sqrt{2 \pi \int_{t}^{t+1}G d\tau}} \exp \left( -\frac{ (x - w^*)^2}{2 \int_{t}^{t+1} G d\tau}  \right) dx \\
&\geq \exp\left(- \frac{C_3}{\sigma (s)} \right) \frac{1}{\sigma(0) \sqrt{2 \pi \| G \|}} \exp \left( -\frac{ (\sqrt{\rho} + \varepsilon)^2}{2  \|G\| } \right) 
\geq \exp\left(- \frac{C_3}{\sigma (s)} \right) \cdot C_4 \cdot \; \because C_4 = \frac{\sqrt{2}}{\sigma(0) \sqrt{\pi \| G \|}}. 
\end{aligned}
\label{lmpf2.7_eq19}    
\end{equation}

Finally, we obtain the lower bound of the transition probability  such that 
\begin{equation}
\begin{aligned}
\delta_t 
&= \inf_{x, y \in \mathbf{R}^n} p(t, x, t+1, y) \bigg\vert_{x = w_t,\; y=w^*} 
= \inf_{x, y \in \mathbf{R}^n} \lim_{\varepsilon \rightarrow 0} \frac{1}{\varepsilon} P_w (|W_{t+1} - w^*| < \varepsilon) \\
&\geq \inf_{x, y \in \mathbf{R}^n} \lim_{\varepsilon \rightarrow 0} \frac{1}{\varepsilon} \cdot C_4 \cdot \exp\left(- \frac{C_3}{\sigma (s)} \right) \cdot  \cdot \varepsilon 
\geq \exp \left(-\frac{C_5}{\sigma(s)}  \right), \quad \because C_5 > C_3 + \sigma(0) \cdot | \ln C_4 |
\label{lmpf2.7_eq20}
\end{aligned}    
\end{equation}
Therefore, if there exists a monotone decreasing function such that $\sigma(s) \geq \frac{C_5}{ \log (t+2)}$ , it satisfies that the convergence condition derived by $\eqref{lmpf2.7_eq04}$ such that
\begin{equation}
\sum_{k=0}^{\infty} \delta_{t+k} = \infty, \quad \forall k \geq 0.    
\end{equation}
It implies that 
\begin{equation}
\overline{\lim_{\tau \rightarrow \infty}} \sup_{w_t, w_{t+\tau} \in \mathbf{R}^n} \| p(t, \bar{w}_t, t + \tau,  w^*) - p(t, w_t, t + \tau,  w^*) \| = 0.
\label{lmpf2.7_eq01}    
\end{equation}
\end{proof}
\subsection{Auxiliary Sub-Functions for Main Algorithm}

\begin{algorithm2e}[H]
    \caption{Avoid Gradient Vanishing and Check the Limit of $\bar{h}_t$}
    \label{algorithm-2}
    \DontPrintSemicolon
    \SetAlgoLined
    \KwData{$h^Q, \; t$}
    \KwResult{Re-quantized $h^Q$, $\sup_{t \geq 0} \bar{h}(t)$}
    $\sup_{t \geq 0} \bar{h}(t) \leftarrow \frac{1}{2} \log_b \left( \frac{n \log (t+2)}{24 \cdot \eta^2 \cdot C} \right)$\;
    \While{$\|h^Q\| > 0$ or $\bar{h}(t) > \sup \bar{h}(t)$}{
        \eIf {$\|h^Q\| = 0$ and $\bar{h}(t) \leq \sup \bar{h}(t)$}
        {
            $\bar{h}(t) \leftarrow \bar{h}(t) + 1$ \tcp*{Increase Resolution of Quantization by 1}
            $Q_p(t) \leftarrow  \eta \cdot b^{\bar{h}(t)}$\;
            $h^Q \leftarrow \frac{1}{Q_p}(Q_p \cdot h_t)^Q$ \tcp*{Re-quantization with updated $Q_p$}
        }
        { 
            $h^Q \leftarrow h^Q$
        }
    }
\end{algorithm2e}

Herein, we provide auxiliary sub-algorithms for the proposed main procedure illustrated as Algorithm $\ref{algorithm-1}$.
The Algorithm $\ref{algorithm-2}$ is a sub-function to avoid vanishing gradient by quantization. The Algorithm $\ref{algorithm-3}$ is a sub-function to limit quantization parameter between $\sup \bar{h}(t)$ and $\inf \bar{h}(t)$.

\begin{algorithm2e}[H]
    \caption{Check the bound of $\bar{h}(t)$}
    \label{algorithm-3}
    \DontPrintSemicolon
    \SetAlgoLined
    \KwData{$\bar{h}(t), \; \sup_{t \geq 0} \bar{h}(t)$}
    \KwResult{Updated $\bar{h}(t)$}
    $\inf \bar{h}(t) \leftarrow -\frac{\beta}{t+2} + \sup_{t \geq 0} \bar{h}(t)$ \;
    \While{$\bar{h}(t) \geq \inf \bar{h}(t)$}{
        \eIf {$\bar{h}(t) < \inf \bar{h}(t)$}
        {
            $\bar{h}(t) \leftarrow \bar{h}(t)+1$ \tcp*{Increase Resolution of Quantization by 1}
        }
        {
            $\bar{h}(t) \leftarrow \bar{h}(t)$
        }
    }
\end{algorithm2e}

\subsection{Hyper-parameters for algorithms}
We set the Hyper-parameters for each algorithms in numerical experiments as follows: 
\subsubsection{SGD}
\begin{itemize}
    \item Directional Derivative 
    \begin{equation}
    h(w_t) = \nabla f(w_t)        
    \end{equation}
    \item Hyper-Parameter 
    $\lambda \in \mathbf{R}(0, 1)$
\end{itemize}

\subsubsection{ADAM}
\begin{itemize}
    \item Directional Derivative 
    \begin{equation}
    h(w_t) = \frac{\sqrt{1 - (\beta_2)^t}}{1 - \beta^t}\cdot \frac{m_t^i}{\sqrt{v_t^i} + \varepsilon}     
    \end{equation}
    \begin{itemize}
        \item First  order Momentum : $m_t^i = \beta_1 m_{t-1}^i + (1-\beta_1)\nabla f(w_t)^i$
        \item Second order Momentum : $v_t^i = \beta_2 v_{t-1}^i + (1-\beta_2)(\nabla f(w_t)^i)^2$
    \end{itemize}
    \item Hyper parameters
    \begin{equation}
    \lambda \in \mathbf{R}(0, 1), \eta_t = 0.001, \beta_1 = 0.9, \beta_2 = 0.999       
    \end{equation}
\end{itemize}

\subsubsection{Proposed Quantization}
For the proposed algorithm, since it is a quantization method, there is not a directional derivation. 
\begin{itemize}
    \item Hyper parameters
    \begin{equation}
    \bar{h}(0) = 2, b=2, \eta=1, \beta=20.0, C=10^{6}       
    \end{equation}
    Therefore, the initial value of the quantization parameter is $Q_p(0) = b^{\bar{h}(0)}$
\end{itemize}

\newpage
\bibliography{_001__main_paper}

\end{document}